\documentclass[twoside,11pt]{article}
\usepackage{jair,theapa, rawfonts}

\usepackage{times}

\usepackage{amssymb}
\usepackage{amsthm}
\usepackage{amsmath}
\usepackage{algorithm}
\usepackage[noend]{algpseudocode}
\usepackage{float}
\usepackage[titletoc,toc,title]{appendix}

\usepackage[titletoc,toc,title]{appendix}
\usepackage{graphicx}
\usepackage{multirow}
\usepackage{ctable}
\usepackage{color}

\theoremstyle{definition}
\newtheorem{defn}{Definition}[section]

\jairheading{1}{2015}{3-17}{6/91}{9/91}

\title{Recursive Feature Generation for Knowledge-based Learning}
\author{\name Lior Friedman \email liorf@cs.technion.ac.il \\
	\name Shaul Markovitch \email shaulm@cs.technion.ac.il \\
	\addr Technion-Israel Institute of Technology\\
	Haifa 32000, Israel
}

\begin{document}
\maketitle
\begin{abstract}
	When humans perform inductive learning, they often enhance the process with background
	knowledge. With the increasing availability of well-formed collaborative knowledge bases, the
	performance of learning algorithms could be significantly enhanced if a way were found to exploit
	these knowledge bases. In this work, we present a novel algorithm for injecting external knowledge
	into induction algorithms using feature generation. Given a feature, the algorithm defines a new
	learning task over its set of values, and uses the knowledge base to solve the constructed learning
	task. The resulting classifier is then used as a new feature for the original problem. We have
	applied our algorithm to the domain of text classification using large semantic knowledge bases. We
	have shown that the generated features significantly improve the performance of existing learning
	algorithms.
\end{abstract}

\section{Introduction}
\label{sec:Intro}
In recent decades, machine learning techniques have become more prevalent in a wide variety of fields. 
Most of these methods rely on the inductive approach: they attempt to locate a hypothesis that is  supported by given labeled examples. These methods have proven themselves successful when the number of examples is sufficient, and a collection of good,
distinguishing features is available.
In many real-world applications, however, the given set of features is insufficient for inducing a high quality classifier \cite{levi2004learning,paulheim2012unsupervisedfull}.

One approach for overcoming this difficulty is to generate new features that are combinations of the given ones.
For example, the LFC algorithm \cite{ragavan1993complex} combines binary features through the use of logical operators such as $\land ,\lnot$.
Another example is the LDMT algorithm \cite{utgo1991linear}, which generates linear combinations of existing features.
Deep Learning methods combine basic and generated features using various activation functions such as sigmoid or softmax.
The FICUS framework \cite{markovitch2002feature} presents a general method for generating features using any given set of constructor functions.

The above approaches are limited in that they merely combine existing features to make the representation more suitable for the 
learning algorithm.  While this approach often suffices, there are many cases where simply combining existing features is not sufficient.
When people perform inductive learning, they usually rely on a vast body of background knowledge to make the process more
effective \cite{mcnamara1996learning}. For example, assume two positive examples of people suffering from some genetic disorder, where the 
value of the country-of-origin feature is Poland and Romania.  Existing induction algorithms, including those generating features 
by combination, will not be able to generalize over these two examples.  Humans, on the other hand, can easily generalize 
and generate a new feature, \emph{is located in Eastern Europe}, based
on their previously established background knowledge.

In this work, we present a novel algorithm that uses a similar approach for enhancing inductive learning with background knowledge through feature generation.
Our method assumes that in addition to the labeled set of example it receives a body of external knowledge represented in relational form.  The algorithm treats feature values as objects and constructs new learning problems, using the background relational knowledge as their features.  The resulting classifiers are then used as new generated features.
For example, in the above very simple example, our algorithm would have used the \emph{continent} and \emph{region} features of a country,
inferred from a geographic knowledge base, to create the new feature that enables us to generalize.  One significant advantage of
using background knowledge in the form of generated features is that it allows us to utilize existing powerful learning algorithms for the induction process.

We have implemented our algorithm in the domain of text classification, using Freebase and YAGO2 as our background knowledge bases, 
and performed an extensive set of experiments to test the effectiveness of our method.  Results show that the use of background knowledge through our methodology significantly improves the performance of existing learning algorithms.

\section{Motivation} \label{motivation}

\begin{figure}[h]
	\centering
	\includegraphics[width=0.9\linewidth]{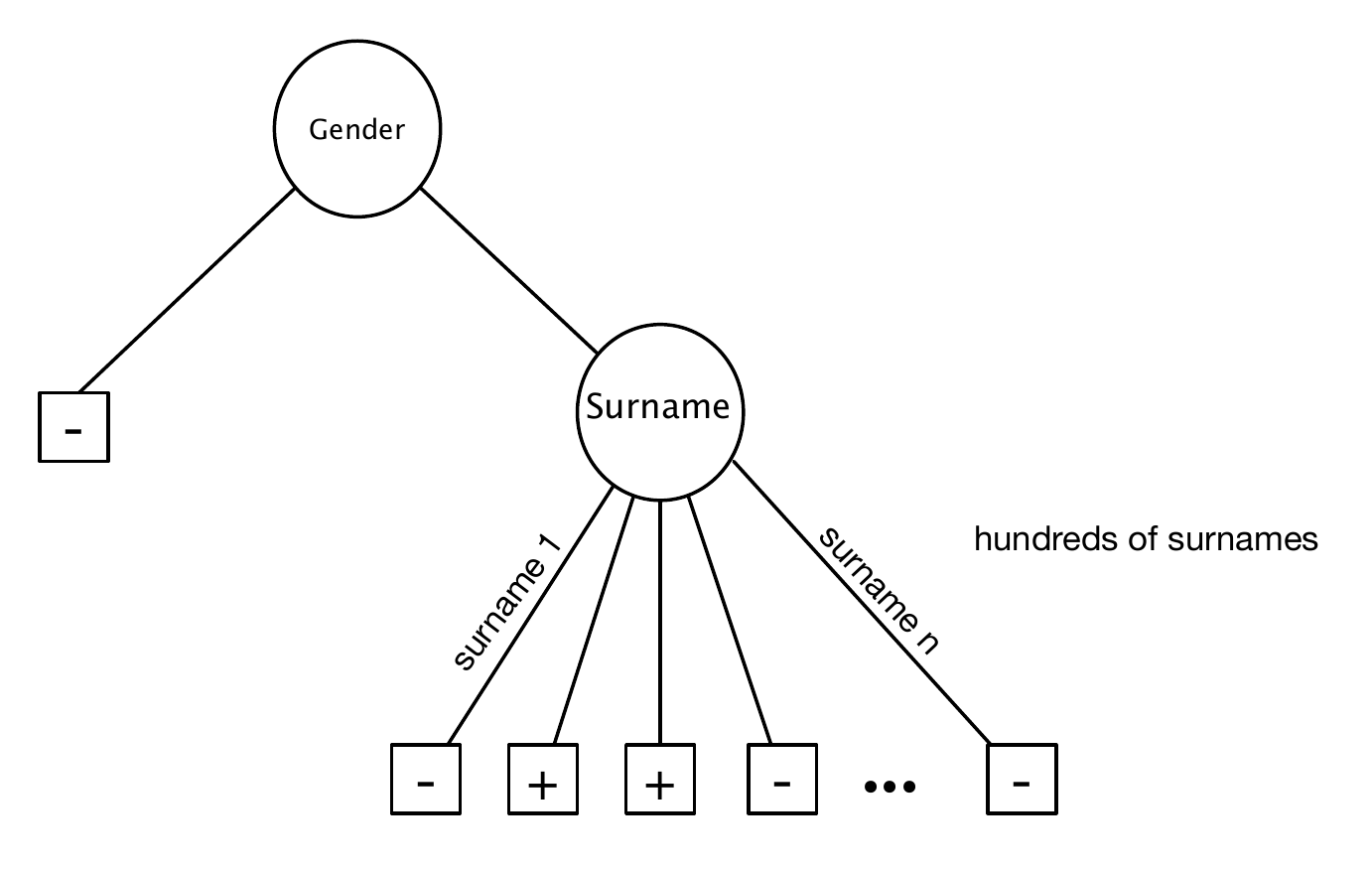}
	\caption{A decision tree for the basic features}
	\label{fig:tree_base}
\end{figure}

Before we delve into the detailed description of our algorithm, we would like to illustrate its main ideas using an example.
Suppose we are attempting to identify people with a high risk of suffering from a genetic disorder. Assume that the target concept to be discovered is that those at risk are women with ancestors originating from desert areas. To identify women at risk, we are given a training sample of sick and healthy people, containing various features, including gender and their full name. We call this learning problem $T_1$.
Assuming we have no additional information, an induction algorithm (a decision tree learner, in this example) would likely produce a result similar to that shown in Figure \ref{fig:tree_base}. While such a classifier will achieve a low training error, the hundreds of seemingly unrelated surnames will cause it to generalize poorly. 

\begin{figure}[h]
	\centering
	\includegraphics[width=0.9\linewidth]{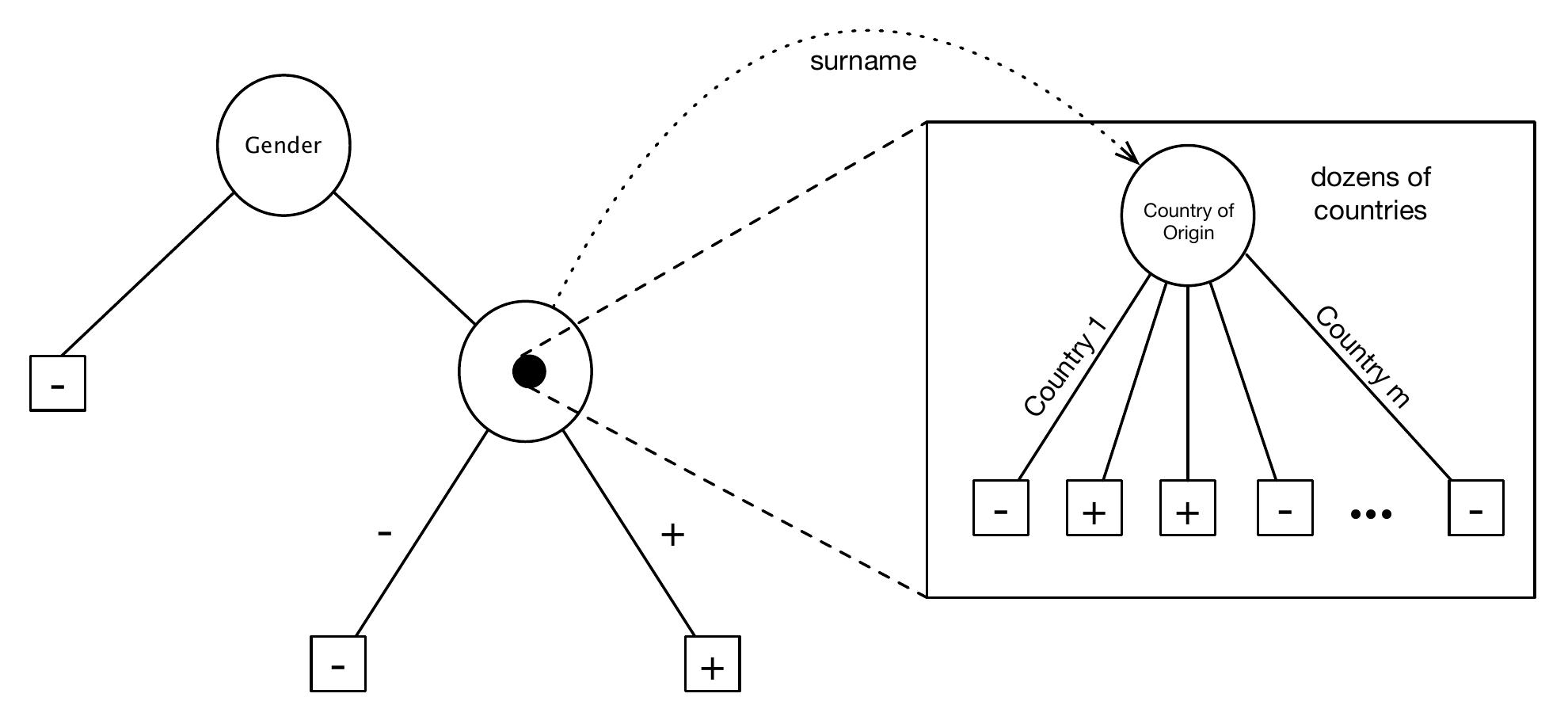}
	\caption{A constructed feature used within a decision tree}
	\label{fig:lvl1_tree}
\end{figure}

The above example illustrates a case where, without additional knowledge, an induction algorithm will yield a very poor result. 
However, if we assume access to a relational knowledge base connecting surnames to common countries of origin, we can begin to apply our knowledge-based feature generation techniques to the problem, as we can move from the domain of surnames to that of countries. 
Our algorithm does so by creating a new learning problem $T_2$. The training objects for learning problem $T_2$ are surnames; surnames of people at risk are labeled as positive. The features for these new objects are extracted from the knowledge base. In this case, we have a single feature: the country of origin.
Solving the above learning problem through an induction algorithm yields a classifier on surnames that distinguishes between surnames of patients with the disease and surnames of healthy individuals. This classifier for $T_2$ can then be used as a binary feature for the original problem $T_1$ by applying it to the feature value of surname. For example, it can be used as a feature in the node corresponding a gender of female in Figure \ref{fig:tree_base}, yielding the tree seen in Figure \ref{fig:lvl1_tree}. 

This new feature gives us a better generalization over the baseline solution, as we now abstract the long list of surnames to a short list of countries. 
This result also allows us to capture previously unseen surnames from those countries. However, this is not a sufficient solution, as we have no way of generalizing on previously unseen countries of origin. 

If, however, we would have recursively applied our method for solving $T_2$, we could have obtained a better generalization.
When learning to classify surnames, our method creates a new learning problem, $T_3$, with countries as its objects. Countries of surnames belonging to people with high risk are labeled as positive. The knowledge base regarding countries is then used to extract features for this new training set.
Applying a standard learning algorithm to $T_3$ will yield a classifier that separates between countries of origin of people at risk and those not at risk. This classifier will do so by looking at the properties of countries, and conclude that countries with high average temperature and low precipitation, the characteristics of desert areas, are associated with people at high risk.

\begin{figure*}[th]
	\centering
	\includegraphics[width=0.9\linewidth,height=0.35\linewidth]{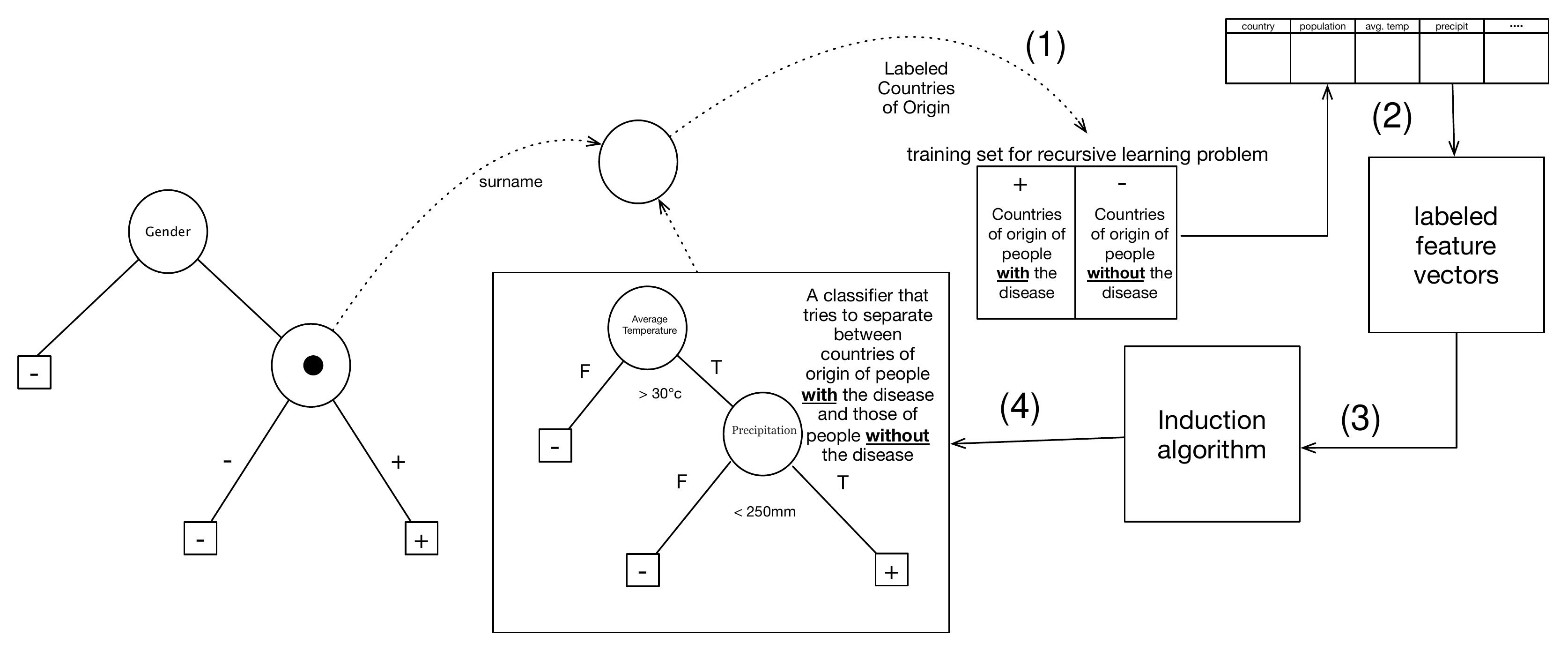}
	\caption{Recursive construction of a learning problem on countries of origin. $(1)$ Creating the objects for the new problem. $(2)$ Creating features using the knowledge base. $(3)$ Applying an induction algorithm. $(4)$ The resulting feature.}
	\label{fig:moving_to_lvl2}
\end{figure*}

The result of this process, depicted in Figure \ref{fig:moving_to_lvl2}, is a new feature for $T_2$, that is, a feature on surnames. This feature is then used to construct a classifier for $T_2$, which is in turn used as a feature for $T_1$, yielding a feature on patients. This new feature for patients will check whether their surname corresponds with a country of origin that has desert-like characteristics. We see that this feature allows us to correctly capture the target concept.


\section{Generating features through recursive induction} \label{formal}

In the following sections, we formally define the feature generation problem, present a solution  in the form of a simple algorithm that generates features by using
relational expansions, and then proceed to describe our main recursive feature generation algorithm.

\subsection{Problem definition}

We begin our discussion with a standard definition of an induction problem. 
Let $O$ be a set of objects. Let $Y=\{0,1\}$ be a set of labels\footnote{We assume binary labels for ease of discussion.}. Let $C:O\rightarrow Y$ be a target concept. Let $S=\{(o_{1},y_{1}),\ldots,(o_{m},y_{m})\}$ be a set of labeled examples such that $o_{i}\in O, y_{i}\in Y, C(o_i)=y_i$. 
Let $F=\{f_{1},\ldots,f_{n}\}$ be a \emph{feature map}, a set consisting of \emph{feature functions} $f_{i}:O\rightarrow I_{i}$ where $I_{i}$ is the image of $f_i$.  This definition implies a training set represented by feature vectors: $S_F=\{ (\langle f_1(o_i),\ldots,f_n(o_i)\rangle, y_i) | (o_i,y_i) \in S\}$. A learning algorithm $L$ takes $S_F$ as inputs, and outputs a classifier $h_{S_F}:O\rightarrow Y$.
\begin{defn}
	Let $L(S,F)=h_{S_F}$ be the classifier given as an output by $L$ given $\langle S,F\rangle$. Assuming $S\sim D$, the generalization error of a learning algorithm $L$ is the probability $Pr(h_{S_F}(x)\neq y)$, where $(x,y)\sim D$.
\end{defn}

\begin{defn}
	A \emph{feature generation algorithm} $A$ is an algorithm that, given $\langle S,F\rangle$, creates a new feature map $F'=\{f'_{1},\ldots,f'_{k}\}, f'_{i}:O\rightarrow I_i$.
\end{defn}

In order to evaluate the output of a feature generation algorithm $A$, we must define its utility. Given $\langle S,F \rangle$, $A$ generates a feature set $F'_A$.
Let $S$ be a training set representative of the true distribution of the target concept.
Given $S$ , a feature set $F$, a generated feature set $F'_A$ and a learning algorithm $L$, the utility of $A$ is $U(A(S,F))=Pr(h_{S_F}(x)\neq y)-Pr(h_{S_{F'_A}}(x)\neq y)$, where $x\in O,y=C(x)$.

Thus, in order for the utility of $A$ to be positive, the generated feature set $F'_A$ must yield a lower generalization error than the original feature map $F$.

In this work, we assume that, in addition to $S_F$, $A$ also has access to a set of binary\footnote{If our relations are not binary, we can use projection to create multiple binary relations instead.} relations ${\cal R}=\{R_{1},\ldots,R_{t}\}, R_j:D_j\times D_{j'}$ representing our knowledge base. For each individual relation $R_j$, its set of departure is marked $D_j$ and its co-domain is denoted as $D_{j'}$. 
\begin{defn}
	A \emph{knowledge-based feature-generation algorithm} $A$ is an algorithm that, given $\langle S,F,{\cal R} \rangle$, creates a new feature map $F_{{\cal R}}=\{f'_{1},\ldots,f'_{k}\}, f'_{i}:O\rightarrow I_i$.
\end{defn}


\subsection{Expansion-based feature generation} \label{shallow_section}

In this section, we present our first method for knowledge-based feature generation that ignores the labels of the original learning problem.
The algorithm extends each feature value by all of the tuples it appears in.
Let $f_i$ be an original feature. Let $R_j:D_j\times D_{j'}$ be a relation such that  $Image(f_i) \subseteq D_j$. We can generate a new feature function $f_{i,j}:O\rightarrow D_{j'}$ by composing $R_j$ onto $f_i$, yielding our new feature function  $f_{i,j}(x)=R_j\circ f_i$.


In the general case, composing $R_j$ onto $F_i$ yields a set of values, meaning $f_{i,j}(x)=\{v\in D_{j'}|(f_i(x),v)\in R_j\}$. 
In our work, we preferred to work on singular values rather than set-based features. To do so, we use aggregation functions.
For the experiments described in this paper, we use  
two aggregation function types: Majority and Any, but in general any other reasonable aggregation function can be used instead.
The Majority aggregator, for example, is defined as follows.  For each value $v\in D_{j'}$, we generate a binary feature function with value 1 only if $v$ is the majority value: 
$Majority^v(X)=1 \iff majority(X)=v$.
The pseudo-code of this algorithm (called   
\emph{Expander-FG}) is listed in Algorithm \ref{code-compete}.


\begin{algorithm}[H]
	\caption{\emph{Expander-FG}}
	\label{code-compete}
	\small
	$\sigma$ - An aggregation function family
	\begin{algorithmic}
		\Function{GenerateFeatures}{$S$,$F$, ${\cal R}$}
		\State $generated=\emptyset$
		\For {$f_i \in F$}
		\For {$R_j \in {\cal R}$ such that $Image(f_i)\subseteq D_{j}$ }
		\If {$R_j$ is a function}
		\State add $\{f_{i,j}=R_j\circ f_i\}$ to $generated$
		\Else \Comment $R_j$ is a relation $R_j:D_j\times D_{j'}$
		\State add $F^\sigma_{i,j}=\{\sigma^v(f_{i,j})|v\in D_{j'}\}$ to $generated$
		\EndIf
		\EndFor
		\EndFor
		\State \Return $generated$ 
		\EndFunction
		
	\end{algorithmic}
\end{algorithm}

\subsection{Recursive feature generation algorithm}
\label{algorithm_section}
One way to extend the \emph{Expander-FG} algorithm described in the previous section is to apply it repeatedly to its own output.
Extending the algorithm in this fashion, however, would yield an exponential increase in the number of generated features.
To that end, we propose an alternative knowledge-based feature generation algorithm. Given an input $\langle S,F,{\cal R} \rangle$, for each feature $f_i\in F, f_i:O\rightarrow I_i$, our algorithm creates a recursive learning problem whose objects are the values of $f_i$, where values associated with positive examples in $S$ are labeled positive. 
Features for this generated problem are created using the relations in ${\cal R}$. Once this new learning problem $\langle S'_i,F_{\cal R}\rangle$ is defined, an learning algorithm is used to induce a classifier $h_i:I_i\rightarrow Y$. Finally, our algorithm outputs a single generated feature for $f_i$, $f'_i(x)=h_i\circ f_i=h_i(f_i(x)), f'_i:O\rightarrow Y$.
Note that during the induction process of the newly created learning problem, we can apply a feature generation algorithm on $\langle S'_i,F_{\cal R},{\cal R} \rangle$. In particular, we can apply the above method recursively to create additional features. We call this algorithm \emph{FEAGURE} (FEAture Generation Using REcursive induction).


Given a feature $f_{i}$, we create a recursive learning problem $\langle S'_i,F_{\cal R} \rangle$. 
Let $v_i(S) = \{v | (o,y) \in S, f_{i}(o)=v\}$ be the set of feature values for $f_i$ in the example set $S$. 
We use $v_i(S)$ as our set of objects. To label each $v \in v_i(S)$, we examine at the labels in the original problem.
If there is a single example $o \in S$ such that $f_i(o)=v$, then the label of $v$ will be the label of $o$. Otherwise, we take the majority label  $label(v)=majority(\{y|(o,y)\in S, f_i(o)=v\})$.

To define our learning problem, we must specify a feature map over $v_i(S)$. Similarly to \emph{Expander-FG}, we use the relations in ${\cal R}$ on the elements in the new training set $S'_i = \{ (v, label(v)) | v \in v_i(S) \}$.
For each $R_j\in {\cal R}$, if it is relevant to the problem domain, meaning that $v_i(S)\subseteq D_j$, we utilize it as a feature by applying it to $v$. If $R_j(v)$ is a set, we use aggregators, as described in the previous section. 
The result of this process is a generated feature map for $S'_i$, denoted as $F_{\cal R}$. 

We now have a new induction problem $\langle S'_i,F_{\cal R} \rangle$.
We can further extend $F_{\cal R}$ by recursively using \emph{FEAGURE}, yielding a new feature map $F'_{\cal R}$. The depth of recursion is controlled by a parameter $d$, that will usually be set according to available learning resources.
We proceed to use a learning algorithm\footnote{For our experiments, we used a decision tree learner, but any induction algorithm can be used.} on $\langle S'_i,F'_{\cal R} \rangle$ in order to train a classifier, giving us $h_i:I_i\rightarrow Y$. We can then use $h_i$ on objects in $S$ as discussed above, giving us a new feature $f'_{i}(x)=h_{i}(f_{i}(x)), f'_{i}:O\rightarrow Y$. 

The full algorithm is listed in Algorithm \ref{code-creating-prob}.
While the \emph{FEAGURE} algorithm can be used as described above, we found it more useful to use it in the context of a divide \& conquer approach, in a manner similar to the induction of decision trees.
In this approach, the set of examples is given as an input to a decision tree induction algorithm. The \emph{FEAGURE} algorithm is applied at each node. 
This allows us to generate features that are locally useful for a subset of examples.
At the end of the process the tree is discarded and the generated features are gathered as the final output.

\begin{algorithm}[H]
	\caption{FEAGURE algorithm}
	\label{code-creating-prob}
	\small
	\begin{algorithmic}
		\Function{GenerateFeatures}{$F$, $S$, ${\cal R}$, $d$}
		\For {$f_i\in F$}
		\State $S'_i,F_{\cal R}$= \Call{CreateNewProblem}{$f_i$,$S$,${\cal R}$,$d$} 
		\State $h_i$= \Call{InductionAlgorithm}{$S'_i,F_{\cal R}$} 
		\State add $f'_i(x)=h_i\circ f_i$ to generated features
		\EndFor
		\State \Return generated features
		\EndFunction

		\Function{CreateNewProblem}{$f_{i}$, $S$, ${\cal R}$, $d$}
		\State $v_i(S) = \{v | (o,y) \in S, f_{i}(o)=v\}$
		\State Let $s(v)=\{o | (o,y)\in S, f_{i}(o)=v\}$
		\State 	$S'_i = \{ (v, \mbox{majority-label}(s(v))) | v \in v_i(S)\}$

		\State $F_{\cal R}=\{R_j(v)| R_j\in{\cal R}, v_i(S)\subseteq D_j\}$
		\If {$d>0$}
		 \State $F_{\cal R}=F_{\cal R}\cup$\Call{GenerateFeatures}{$F_{\cal R}, S'_i,  {\cal R}$, $d-1$}
		\EndIf
		\State \Return $S'_i, F_{\cal R}$ 
		\EndFunction
		
	\end{algorithmic}
\end{algorithm}

\subsection{Finding Locally Improving Features} \label{tree_usage}

Some generated features may prove very useful for separating only a subset of the training data but
may be difficult to identify in the context of the full training data. In the motivating example in Section \ref{motivation}, for instance, examples of male individuals (who are not at risk) are irrelevant to the target
concept, and thus mask the usefulness of the candidates for feature generation.
In this subsection, we present an extension of our FEAGURE algorithm that evaluates features in
local contexts using the divide \& conquer approach. This algorithm uses a decision tree induction
method as the basis of its divide \& conquer process. At each node, the FEAGURE algorithm is
applied to the given set of features, yielding a set of generated features. Out of the expanded (base
and generated) feature set, the feature with the highest information gain measure \cite{quinlan1986}
 is
selected, and the training set is split based on the values of that feature. This feature may or may not
be one of the generated features. We continue to apply this approach recursively to the examples in
each child node, using the expanded feature set as a baseline. Once a stopping criterion has been
reached, the generated features at each node are gathered as the final output. In our case, we stop
the process if the training set is too small or if all examples have the same label. The decision tree
is then discarded.
In addition to generating features that operate within localized contexts of the original induction
problem, our approach offers several advantages for feature generation:
\begin{enumerate}
	\item Orthogonality: Because all examples with the same value for a given feature are grouped
	together, any further splits must make use of different features. Due to this and the fact that
	the features selected in each step have high IG, features chosen later in the process will be
	mostly orthogonal to previously chosen features. This results in a larger variety of features
	overall. The feature chosen as a split effectively prunes the search tree of possible features
	and forces later splits to rely on other features and thus different domains.
	
	\item Interpretability: Looking at the features used at each splitting point gives us an intuitive
	understanding of the resulting subsets. Because of this, we can more easily understand why
	certain features were picked over others, which domains are no longer relevant, and so on.
	
	\item Iterative construction: The divide \& conquer approach allows for an iterative search process,
	which can be interrupted if a sufficient number of features were generated, or when the
	remaining training set is no longer sufficiently representative for drawing meaningful conclusions.
\end{enumerate}

The above advantages give us a strong incentive to utilize this approach when attempting to generate features using \emph{FEAGURE}. 
We call this new algorithm \emph{Deep-FEAGURE}, as it goes into increasingly deeper local contexts. 
Pseudocode for this method is shown in Algorithm \ref{code-tree-thing}.
Through the use of a divide \& conquer approach, we can better identify strong, locally useful features that the \emph{FEAGURE} algorithm may have difficulty generating otherwise. In our experiments, we used this approach to generate features.

\begin{algorithm}[H]
	\caption{Deep FEAGURE- Divide \& conquer feature generation}
	\label{code-tree-thing}
	\small
	minSize: minimal size of a node.
	
	generatedFeatures: A global list of all generated features.
	
	SelectFeature: Method that selects a single feature, such as highest information gain.
	
	\begin{algorithmic}
		\Function{DeepFEAGURE}{$S$, $F$, ${\cal R}$, $d$}
		\If {all examples in $S$ are of same class $c$}
		\State 
		\Return leaf($c$)
		\EndIf
		\If {$|S|<$minSize}
		\State 
		\Return leaf(majority class of $S$)
		\EndIf
		\State localGeneratedFeatures=\Call{FEAGURE}{$S$, $F$, ${\cal R}$, $d$}
		\State add localGeneratedFeatures to generatedFeatures
		\State $f=$ \Call{SelectFeature}{$S$, $F\cup$localGeneratedFeatures}
		\State $children=\emptyset$
		\For{$v\in Domain(f)$}
		\State $S(f)=\{(o,y)\in S|f(o)=v\}$
		\State subTree= \Call{DeepFEAGURE}{$S(f)$, $F\cup$localGeneratedFeatures, ${\cal R}$, $d$}
		\State add subTree to $children$
		\EndFor
		\State \Return $children$
		\EndFunction
		
	\end{algorithmic}
\end{algorithm}

\section{Empirical evaluation}
We have applied our feature generation algorithm to the domain of text classification.

\subsection{Application of \emph{FEAGURE} to Text Classification} \label{text-feagure}



The text classification problem is defined by a set of texts $O$ labeled by a set of categories $Y$ \footnote{We can assume $Y=\{0,1\}$ for ease of analysis.}
such that we create $S=\{(o_i,y_i)|o_i\in O, y_i\in Y\}$. Given $S$, the learning problem is to find a hypothesis $h:O\rightarrow Y$ that minimizes generalization error over all possible texts of the given categories. To measure this error, a testing set is used as an approximation.

In recent years, we have seen the rise of Semantic Linked Data as a powerful semantic knowledge base for text-based entities, with large databases such as Google Knowledge Graph \cite{pelikanova2014google}, Wikidata \cite{vrandevcic2014wikidata} and YAGO2 \shortcite{hoffart2013yago2} becoming common. 
These knowledge bases represent semantic knowledge through the use of relations, mostly represented by triplets of various schema such as RDF, or in structures such as OWL and XML. These structures conform to relationships between entities such as ``born in" (hyponyms), as well as type information (hypernyms).

To use \emph{FEAGURE} for text classification, we use words as binary features and Freebase and YAGO2 as our semantic knowledge bases.
YAGO2 \cite{hoffart2013yago2} is a large general knowledge base extracted automatically from Wikipedia, WordNet and GeoNames.
YAGO2 contains over 10 million entities and 124 million relational facts, mostly dealing with individuals, countries and events.
Freebase \cite{bollacker2008freebase} has been described as ``a massive, collaboratively edited database of cross-linked data." Freebase is constructed as a combination of data harvested from databases and data contributed by users. The result is a massive knowledge base containing 1.9 billion facts. 

To apply our approach to the domain of text classification, we perform a few minor adjustments to the \emph{FEAGURE} algorithm:
\begin{enumerate}
	\item To enable linkage between the basic features and the semantic knowledge bases, we use entity linking software \cite{hoffart2011robust,milne2013open} to transform these words into semantically meaningful entities.
	\item Once we have created a new classifier $h_i$, we cannot simply compose it on $f_i$, since every example might contain multiple entities. To that end,  we apply $h_i$ on each entity and take the majority vote.
	\item Since our features are binary, we use the entities extracted from the text as the set of values $v_i(S)$. We split $v_i(S)$ into several subsets according to relation domains and apply the \emph{FEAGURE} algorithm independently to each domain. 
\end{enumerate}

\subsection{Methodology}

We evaluated our performance using a total of 101 datasets from two dataset collections:

\textbf{TechTC-100} \cite{gabrilovich2004text} is a collection of 100 different binary text classification problems of varying difficulty, extracted from the Open Dictionary project.
We used the training and testing sets defined in the original paper. 
As our knowledge base for this task, we used YAGO2. 
For entity extraction, we used AIDA \cite{hoffart2011robust}, a framework for entity detection and disambiguation. 




\textbf{OHSUMED} \cite{hersh1994ohsumed} is a large dataset of medical abstracts from the MeSH categories of the year 1991. 
First, we took the first 20,000 documents, similarly to \citeauthor{joachims1998text} \shortcite{joachims1998text}.
We limited the texts further to medical documents that contain only a title. 
Due to the relatively sparse size of most MeSH categories, we only used the two with the most documents, C1 and C20. 
The result is a dataset of 850 documents of each category, for a total of 1700 documents.
We used ten-fold cross-validation to evaluate this dataset.
Since the YAGO2 knowledge base does not contain many medical relations, we used Freebase instead. We used the same data dump used by \citeauthor{bast2014easy} \shortcite{bast2014easy}. 

In our experiments, we generated features using the \emph{FEAGURE} algorithm. We then proceeded to use these new features alongside three learning algorithms: SVM \cite{cortes1995support}, K-NN \cite{fix1951discriminatory} and CART \cite{breiman1984classification}.

We compared the performance of a learning algorithm with the generated features to the baseline of the same induction algorithm without the constructed features.
In addition, since we could not obtain the code of competitive approaches for relation-based feature generation (such as FeGeLOD, SGLR), we instead compared our algorithm to \emph{Expander-FG}, which we believe to be indicative of the performance of these unsupervised approaches.





\subsection{Results}

Table \ref{table:acc} shows average accuracies across all 10 folds for OHSUMED, as well as the average accuracies for all 100 datasets in techTC-100. When the advantage of our method over the baseline was found to be significant using a pairwise t-test (with $p<0.05$), we marked the $p$-value. Best results are marked in bold.
For the TechTC-100 dataset, \emph{FEAGURE} shows a significant improvement over the baseline approach. 
Of particular note are the results for KNN and SVM, where the two-level activation of \emph{FEAGURE} (d=2) shows statistically significant improvement over \emph{Expander-FG} as well as the baseline accuracy ($p < 0.05$). 
One notable exception to our good results is the poor performance of K-NN for the OHSUMED dataset. This is likely due to the sensitivity of K-NN to the increase in dimension. 
For SVM as the external classifier, the \emph{FEAGURE} algorithm showed an improvement in accuracy for 87 of 100 datasets for $d=1$, and 91 datasets for $d=2$.
Using a Friedman test \cite{friedman1937use}, we see a significant improvement ($p<0.001$) over the baseline.


\begin{table*}[th!]
	\centering
	\caption{Average accuracy over all datasets. The columns specify feature generation approach, with baseline being no feature generation. The rows specify the induction algorithm used on the generated features for evaluation.
		Results marked with * are significant with $p<0.001$.}
	\label{table:acc}
	\begin{tabular}{|l | l || l | l | l| l|}
		\hline
		Dataset & Classifier & Baseline   & Expander-FG & FEAGURE(d=1)   & FEAGURE(d=2)    \\ \hline
		\multirow{3}{*}{OHSUMED} & KNN  & \textbf{0.777} & 0.756 & 0.769   & 0.75 \\ \cline{2-6}
		& SVM  & 0.797 & 0.804   & 0.816 ($p<0.05$)    & \textbf{0.819 ($p<0.05$)} \\ \cline{2-6}
		
		& CART  & 0.806 & 0.814   & 0.809    & \textbf{0.829 ($p<0.05$)} \\
		
		\specialrule{.15em}{.05em}{.01em} 
		
		\multirow{3}{*}{TechTC-100} & KNN & 0.531 & 0.702* & 0.772* & \textbf{0.775*}  \\ \cline{2-6}
		& SVM  & 0.739 & 0.782*    & 0.796*    & \textbf{0.807*} \\ \cline{2-6}
		
		& CART  & 0.81 & 0.815   & 0.814   & \textbf{0.825 ($p<0.05$)}  \\
		
		\specialrule{.15em}{.05em}{.01em}
		
		
		
		
		
		
		
		
	\end{tabular}
\end{table*}

\begin{figure}
	\centering
	\begin{minipage}{0.45\textwidth}
		\centering
		\includegraphics[width=1.2\textwidth]{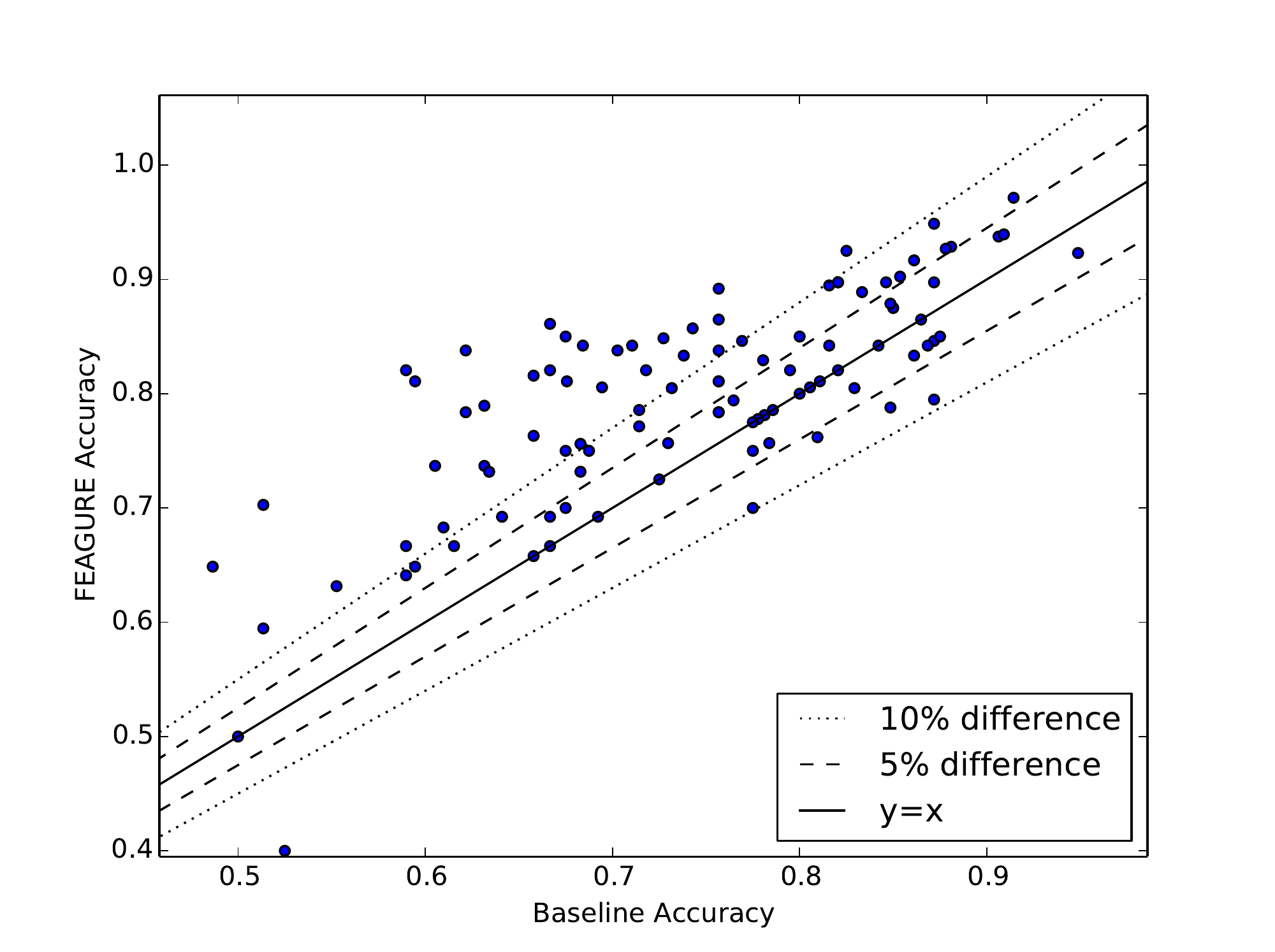} 
		\caption{Accuracy of
			baseline approach compared to one-level activation of \emph{FEAGURE} (SVM). Each point represents a dataset. The dotted lines represent a 5 and 10 percent difference in accuracy}
		\label{fig:svm_base_lvl1}
	\end{minipage}\hfill
	\begin{minipage}{0.45\textwidth}
		\centering
		\includegraphics[width=1.2\textwidth]{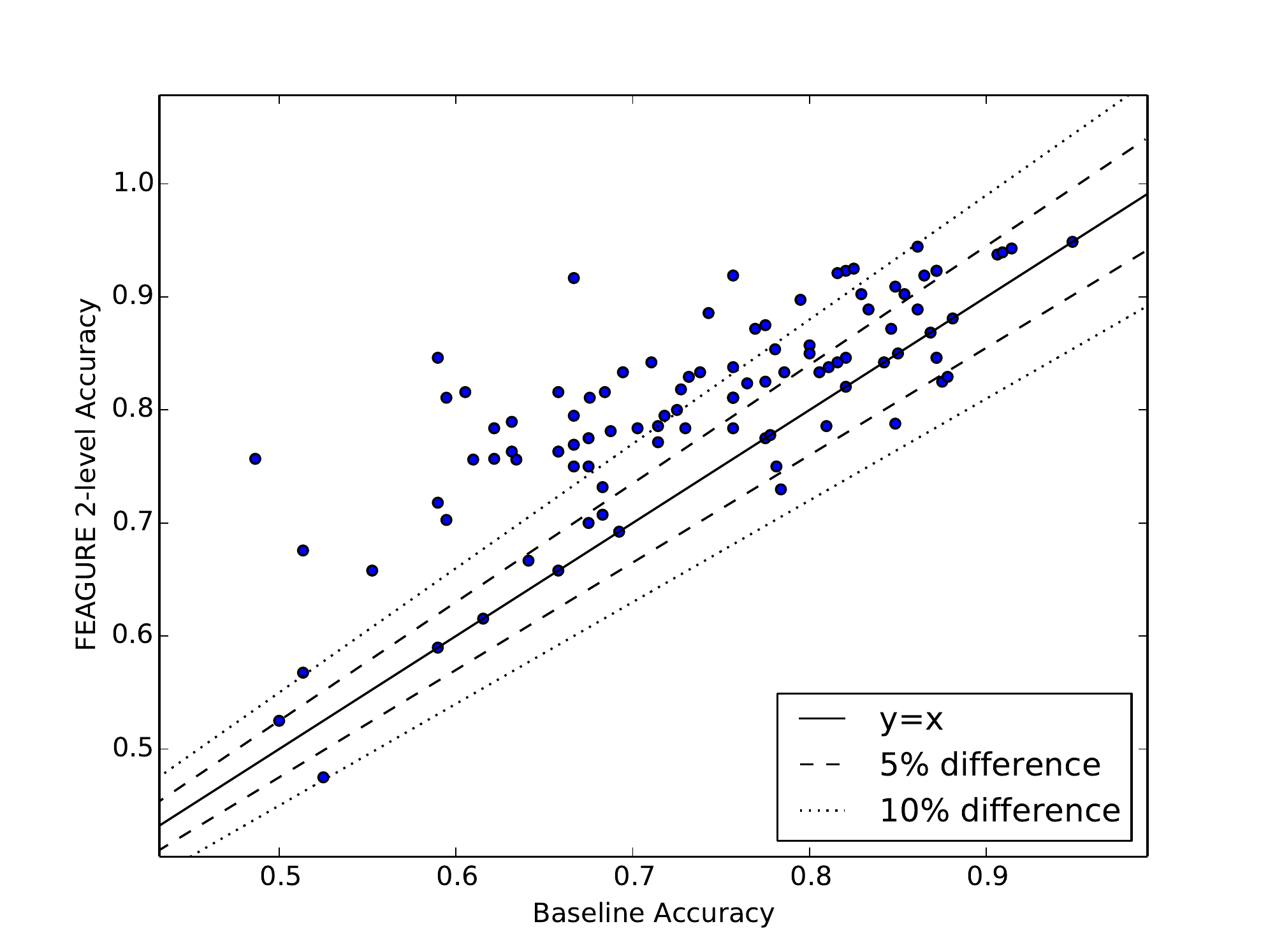} 
		\caption{Accuracy of
			baseline approach compared to two-level activation of \emph{FEAGURE} (SVM). Each point represents a dataset. The dotted lines represent a 5 and 10 percent difference in accuracy}
		\label{fig:svm_base_lvl2}
	\end{minipage}
\end{figure}

Figures \ref{fig:svm_base_lvl1} and \ref{fig:svm_base_lvl2} show the accuracies for datasets in techTC-100 using a SVM classifier. The x-axis represents the baseline accuracy without feature generation, and the y-axis represents the accuracy using our new feature set generated using \emph{FEAGURE}. Therefore, any dataset that falls above the $y=x$ line marks an improvement in accuracy. 
The results show a strong trend of improvement, with high ($>10\%$) improvement being common.
We see that for 8 of the datasets,
there is a degradation in accuracy. This can be a result of mistakes in the entity extraction and
linking process.

\begin{figure}
	\centering
	\includegraphics[width=0.8\linewidth]{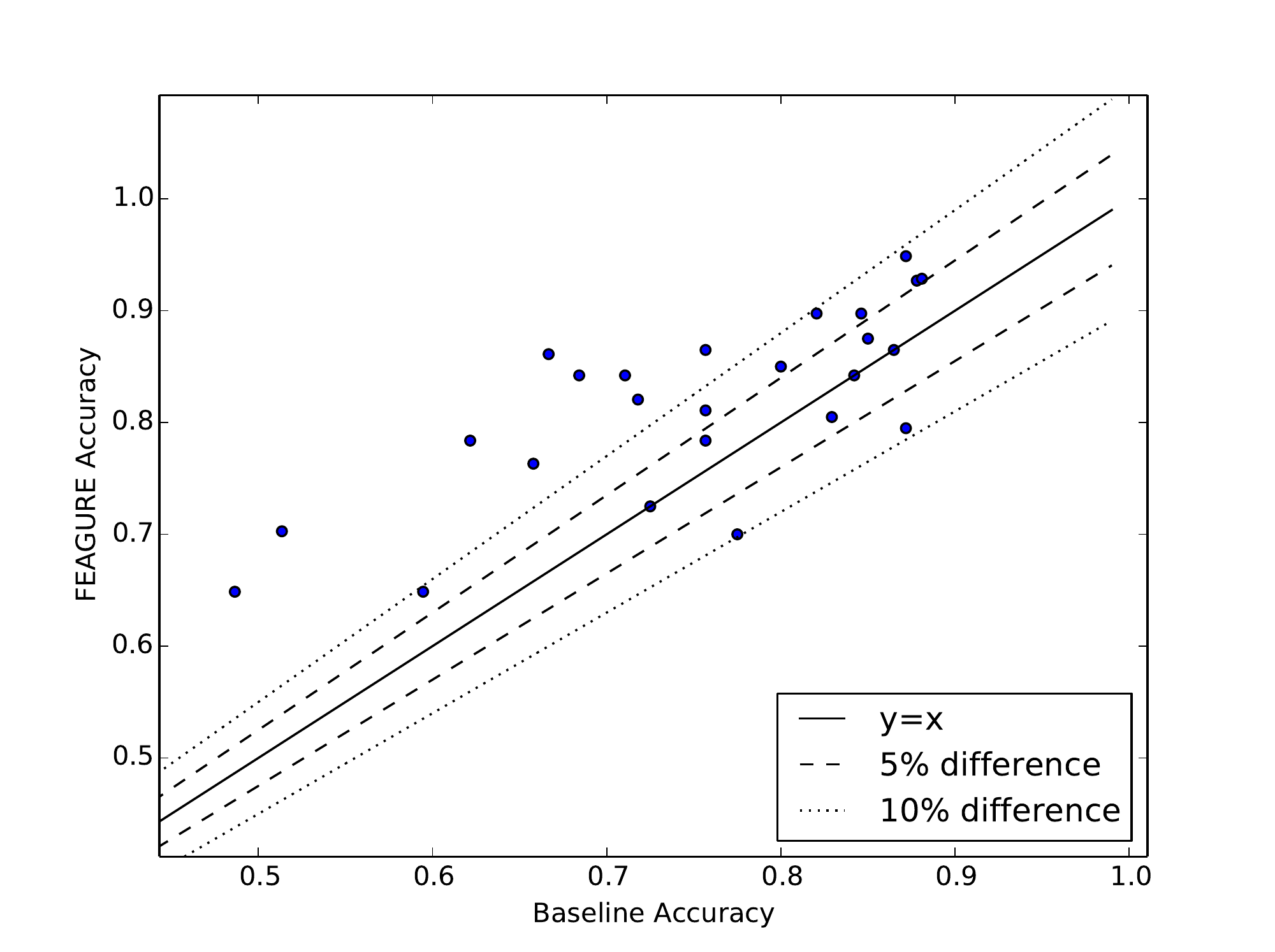}
	\caption{Accuracy of the
		baseline approach compared to single-level activation of \emph{FEAGURE} (SVM). Displayed are the 25 hardest datasets (those with the lowest MAA)}
	\label{fig:25best}
\end{figure}

In their paper on TechTC-100, \citeA{gabrilovich2004text} define a metric called Maximal Achievable Accuracy (MAA). This criterion attempts to assess the difficulty of the induction problem by the maximal ten-fold accuracy over three very different induction algorithms (SVM, K-NN and CART).
Figure \ref{fig:25best} shows the performance of \emph{FEAGURE} on the 25 hardest datasets in TechTC-100, in terms of the MAA criterion. We call this dataset collection ``TechTC-25MAA." 
Table \ref{table:acc_maa} shows the accuracies for ``TechTC-25MAA."
These results show a much more pronounced increase in accuracy, and illustrate that we can, in general, rely on \emph{FEAGURE} to yield positive features for difficult classification problems.

\begin{table}[!h]
	\centering
	\caption{Average accuracy over the 25 hardest datasets in terms of MAA. The columns specify the feature generation approach, with the baseline being no feature generation. The rows specify the induction algorithm used on the generated features for evaluation. Best results are marked in bold.}
	\label{table:acc_maa}
	\begin{tabular}{|l | l || l | l | l| l|}
		\hline
		Dataset & Classifier & Baseline   & Expander-FG & FEAGURE   & FEAGURE 2-level    \\ \hline
		
		\multirow{3}{*}{TechTC-25MAA} & KNN & 0.524 & 0.723 ($p<0.001$) & \textbf{0.803 ($p<0.001$)} & 0.795 ($p<0.001$)  \\ \cline{2-6}
		
		& SVM  & 0.751 & 0.815 ($p<0.001$)    & 0.817 ($p<0.001$)    & \textbf{0.829 ($p<0.001$)} \\ \cline{2-6}
		
		& CART  & 0.82 & 0.839   & 0.837   & \textbf{0.849 ($p<0.05$)}  \\
		
		\specialrule{.15em}{.05em}{.01em}
		
		\multirow{3}{*}{TechTC-100} & KNN & 0.531 & 0.702 ($p<0.001$) & 0.772 ($p<0.001$) & \textbf{0.775 ($p<0.001$)}  \\ \cline{2-6}
		& SVM  & 0.739 & 0.782 ($p<0.001$)    & 0.796 ($p<0.001$)    & \textbf{0.807 ($p<0.001$)} \\ \cline{2-6}
		
		& CART  & 0.81 & 0.815   & 0.814   & \textbf{0.825 ($p<0.05$)}  \\
		
		\specialrule{.15em}{.05em}{.01em}
		
	\end{tabular}
\end{table}

As we have discussed in section \ref{algorithm_section}, \emph{FEAGURE} creates a generic learning problem as part of its execution. For our main results we learned a decision tree classifier for this new induction problem.
We also tested the effects of using K-NN and SVM classifiers instead. 
This choice is orthogonal to that of the learning algorithm used to evaluate the generated features.
Our experiments showed that in general, replacing the internal tree induction algorithm lowers the achievable accuracy of the resulting feature map.
The only exception to this trend is the case of an external K-NN classifier for the OHSUMED dataset. In this case, an internal RBF-SVM induction algorithm yields an average accuracy of $0.795$ (across ten folds), a significant ($p<0.05$) improvement over the baseline. 

\subsection{Quantitative Analysis}


To better understand the behavior of the \emph{Deep-FEAGURE} algorithm, we measured its output and performed several aggregations over it.
We first look at the average number of features considered by \emph{Deep-FEAGURE} at every node: for TechTC-100, out of an average of 36.7 partitions by type that are considered for each feature, an average of 7.3 are expanded into new features by \emph{FEAGURE}. The rest of the generated problems are discarded due to the filtering criteria mentioned in Section \ref{code-tree-thing}.
We note that there are 46 available relations, and thus we can expect to look at no more than 46 features (if a relation does not apply to a sub-problem, it is not counted in this average). 

\begin{figure}[h!]
	\centering
	\includegraphics[scale=0.4]{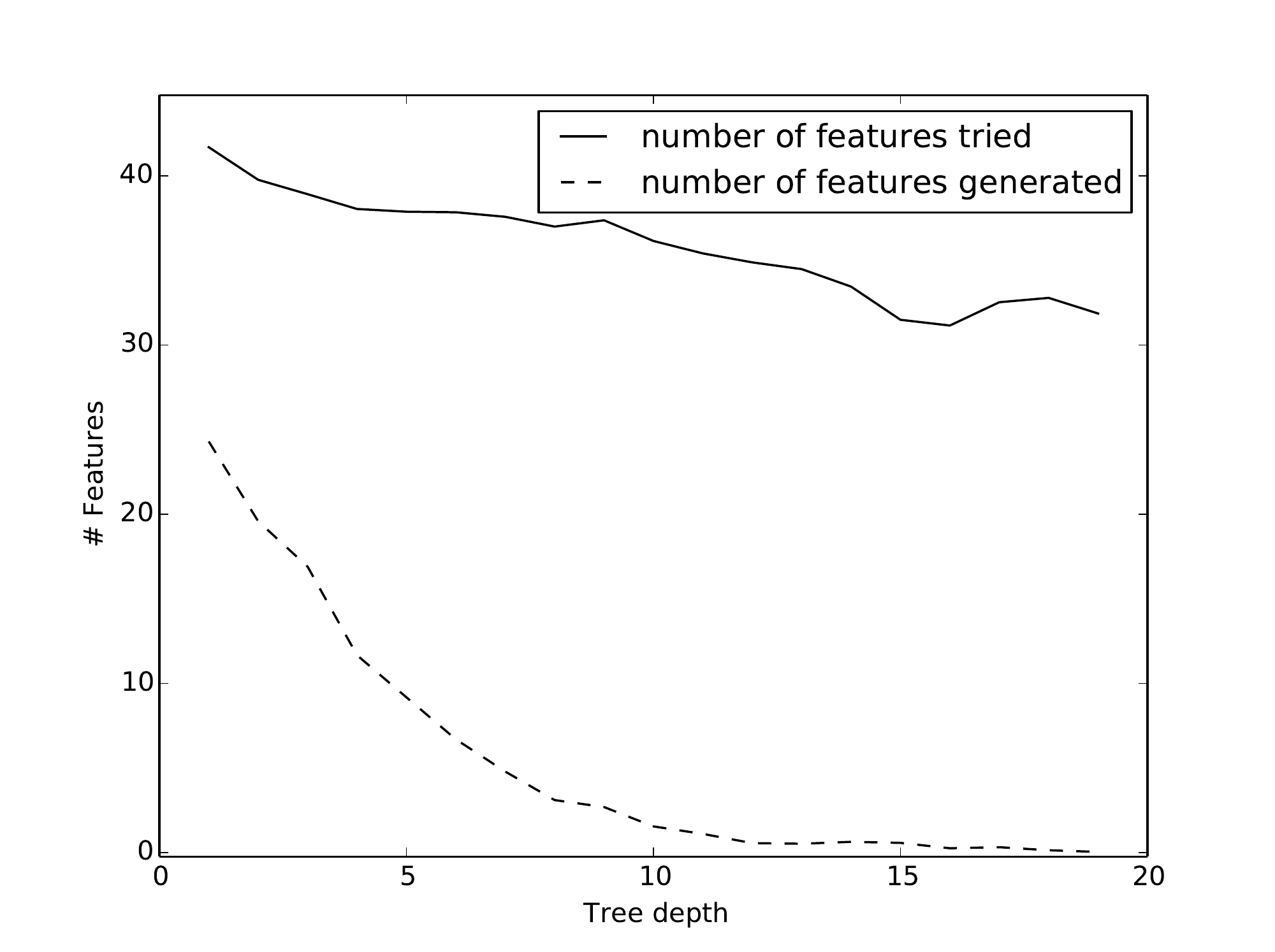}
	\caption{Average number of features tried vs. average number of features generated per depth.}
	\label{fig:features_per_depth}
\end{figure}

When the depth of the search tree increases, the number of generated features decreases rapidly, as shown in Figure \ref{fig:features_per_depth}.
This is unsurprising, as we know that increased depth will cause generated problems to have a smaller training set, increasing the likelihood of those features to be filtered out. Additionally, we see that the number of available features decreases in a roughly linear manner with depth. This is again unsurprising, as features deeper in the tree cannot easily make use of the same relation multiple times due to the orthogonality trait discussed in Section \ref{tree_usage}.

\begin{figure}[h!]
	\centering
	\includegraphics[scale=0.4]{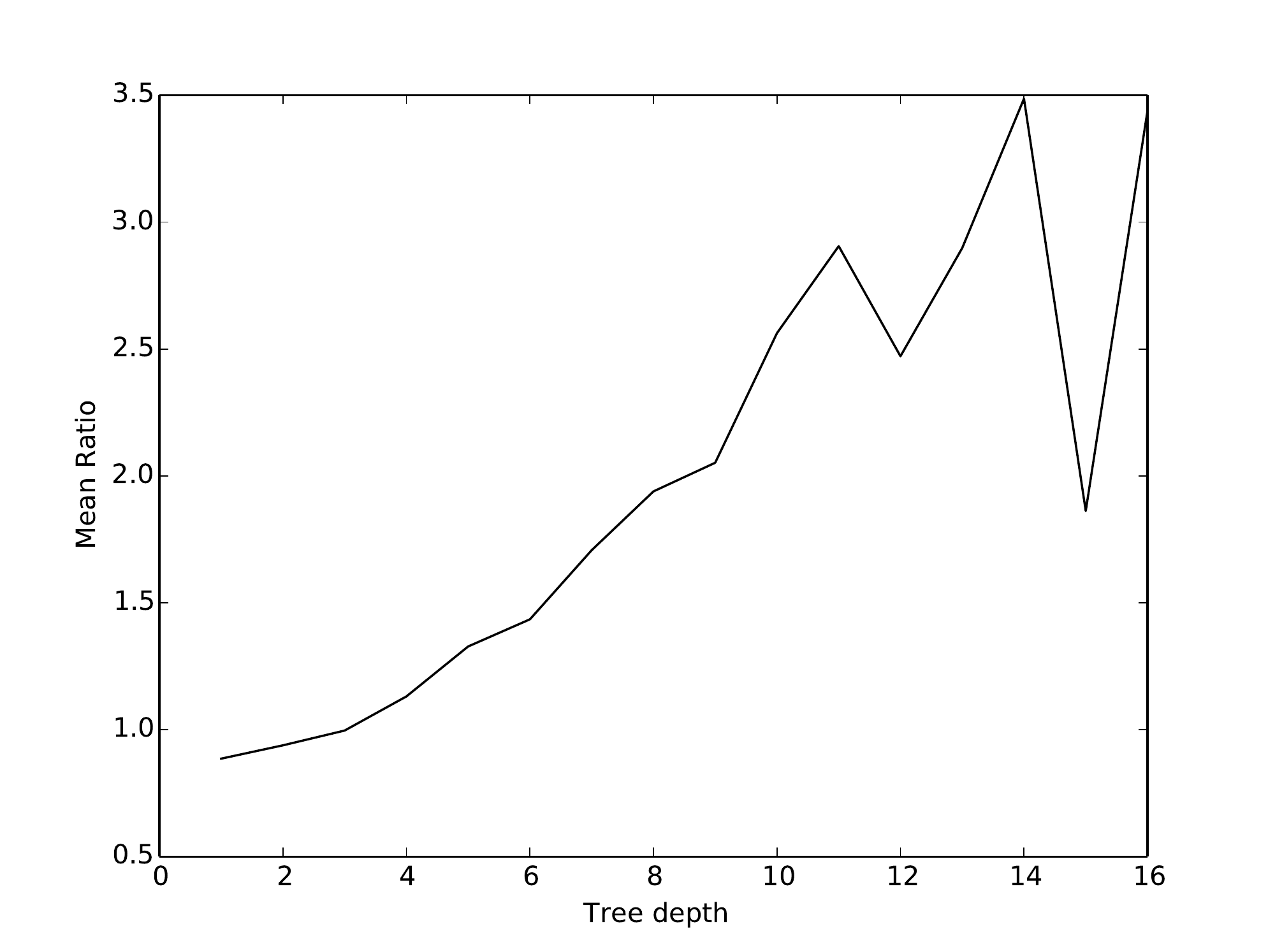}
	\caption{Mean size ratio of newly created recursive problem compared to the original learning problem size (new problem training set size divided by original problem training set size).}
	\label{fig:problem_ratio}
\end{figure}

Let us now compare the relative size of recursive problems to the existing induction problems. Figure \ref{fig:problem_ratio} shows the ratio between the number of examples (size) in the  newly constructed induction problem and the size of the original learning problem at various depths of the divide \& conquer search process. 
As depth in the search tree increases, the size ratio increases as well. This is somewhat surprising, as an increase in depth means a smaller induction problem, and thus we would expect a similar size ratio to be maintained. Even more unexpected, however, is that the average \textbf{recursive} problem size increases with problem depth. Intuitively, we would have expected the recursive problem size to decrease as problem depth increases. However, as we search smaller problems deeper down the search process, the relations that result in a smaller or roughly similar sized recursive induction problem have already been used, and thus relations that cause a larger size ratio are required.

We also note that these recursive problems do not maintain the same label balance as the original learning problems: one label set is much larger. This is expected, as the divide \& conquer search strategy aims to use the label set as a basis for separation. We may therefore wish to make use of various known strategies to reduce the impact of label imbalance in induction problems when using \emph{Deep-FEAGURE}.

\begin{figure}[h!]
	\centering
	\includegraphics[scale=0.4]{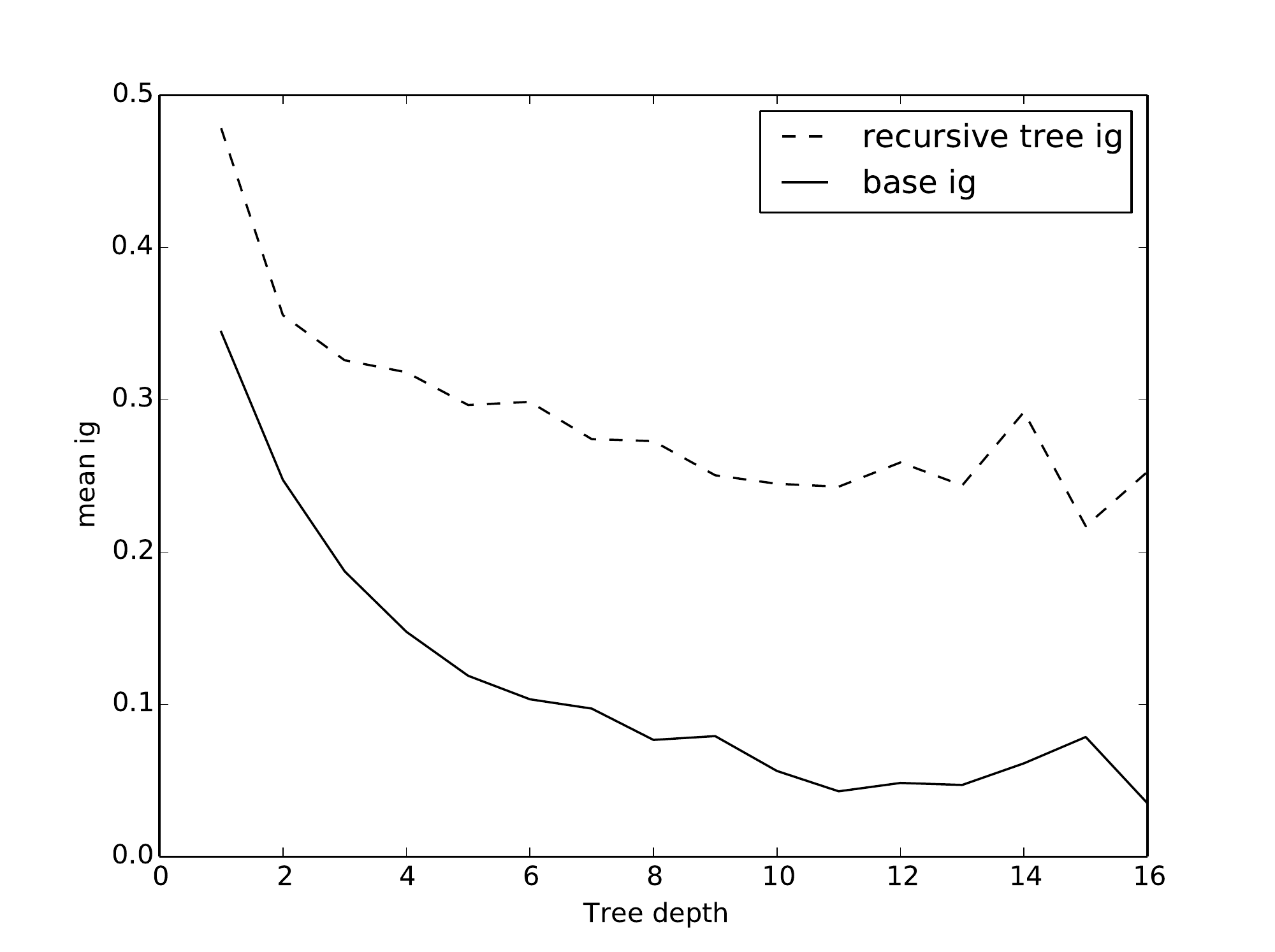}
	\caption{Mean information gain of generated features compared to the best information gain for the original learning problem.}
	\label{fig:local_ig_depth}
\end{figure}

Finally, we look at the local information gain of the newly created features. In Figure \ref{fig:local_ig_depth}, we see the mean information gain of generated features per depth, compared to the best information gain achieved by a non-recursive feature for that depth. We see that our recursive features tend to have a much higher information gain, especially in the beginning of the search process. We see a decrease in both measures as depth increases, because it is then more difficult to find distinguishing features.

\subsection{\emph{FEAGURE} Demonstration}

\begin{figure}[!h]
	\centering
	\includegraphics[width=\linewidth]{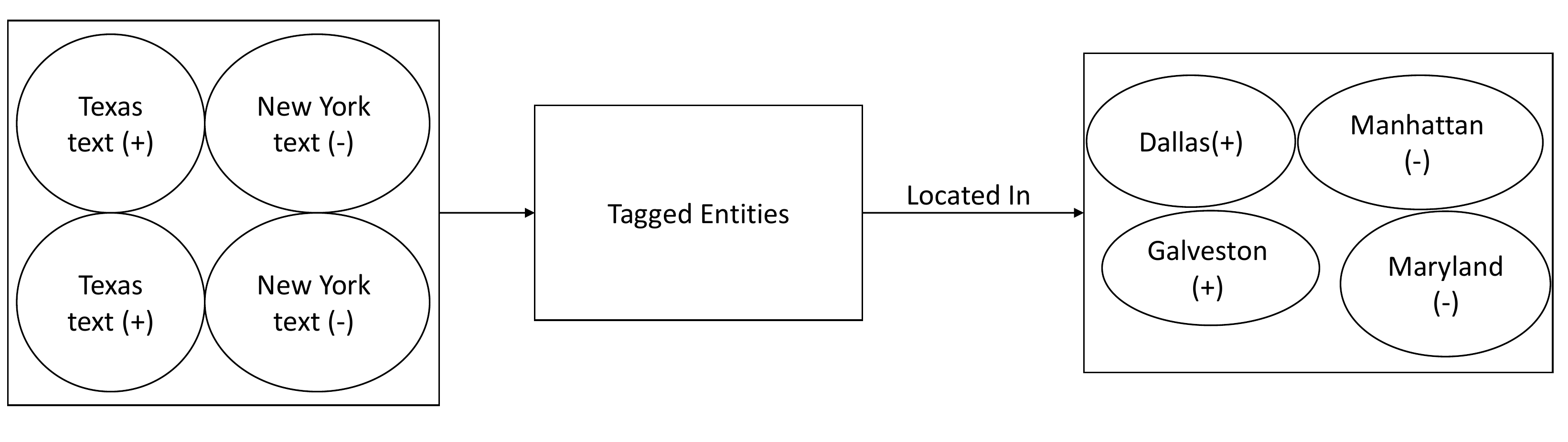}
	\caption{Entities are extracted from the text, and entities in the ``Located in" relation are used as labeled objects.}
	\label{fig:figure_rec3_example}
\end{figure}

\begin{figure}[!h]
	\centering
	\includegraphics[width=\linewidth]{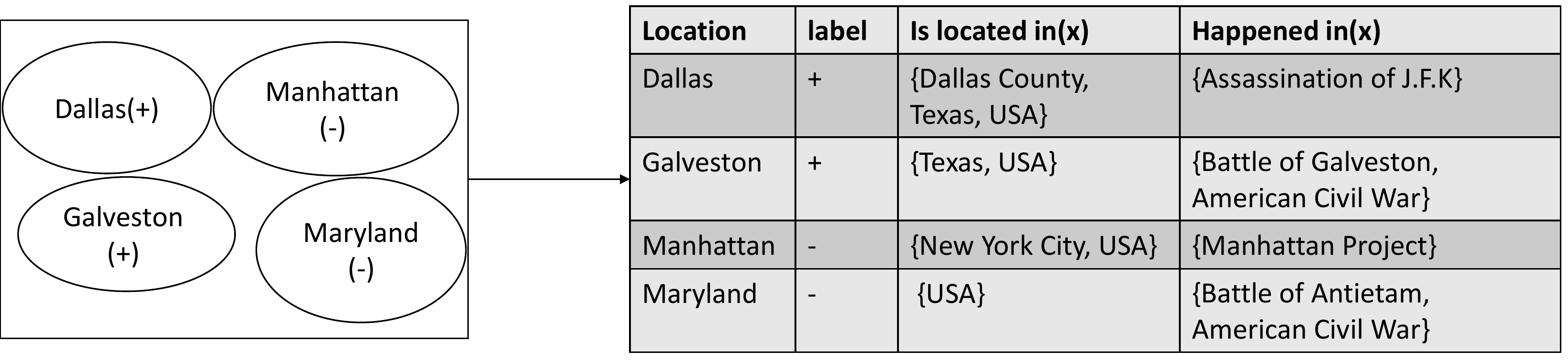}
	\caption{Construction of a recursive learning problem based on the ``Located in" relation. Applicable relations are used to create a feature set for the newly constructed example set.}
	\label{fig:figure_rec3_problem}
\end{figure}

To demonstrate \emph{FEAGURE}, We selected one problem from TechTC-100. In this example, texts refer either to locations in and around Texas, or to locations in and around New York. The extracted entities are locations, with the ``Located in" relation as our domain (Figure \ref{fig:figure_rec3_example}). Applicable relations are used to then create a new induction problem. \emph{FEAGURE} uses the ``Located in" and ``Happened in" relations as features for this problem, as shown in Figure \ref{fig:figure_rec3_problem}. 

\begin{figure}[!h]
	\centering
	\includegraphics[width=\linewidth]{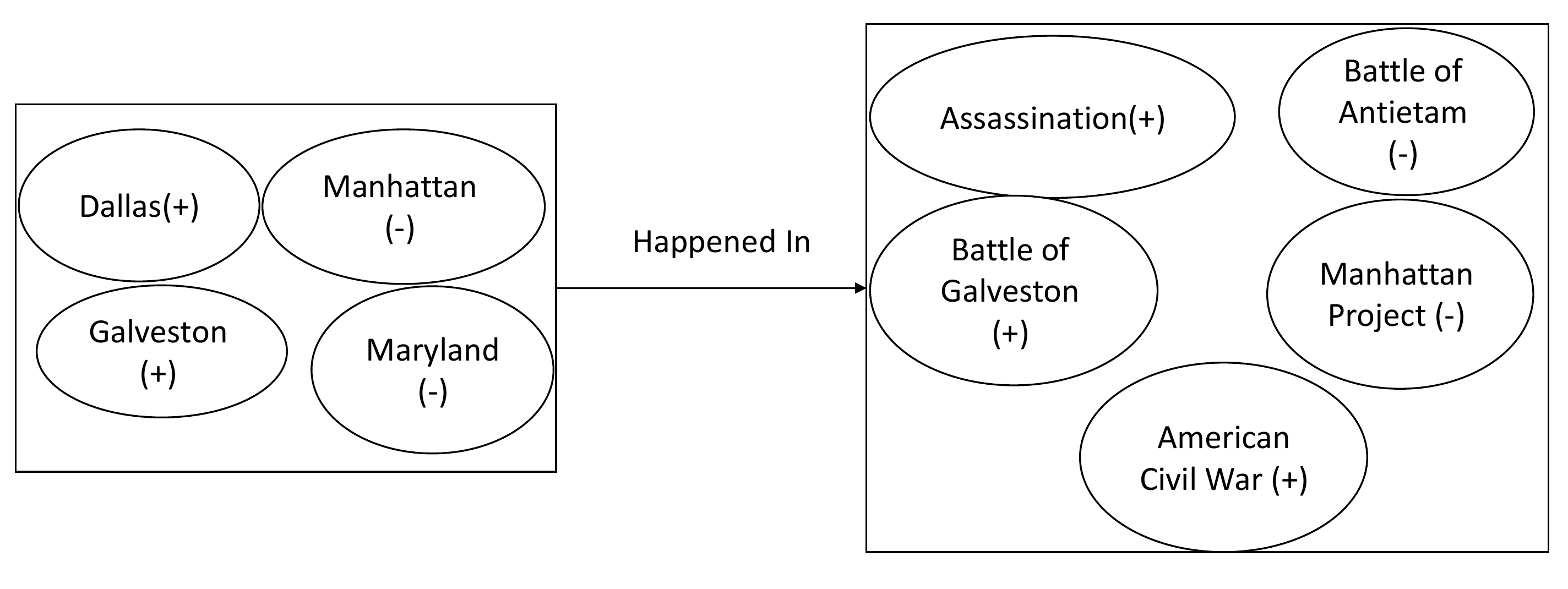}
	\caption{Construction of a second level recursive learning problem based on the ``Happened in" relation. Feature values are treated as objects and labeled according to the labels of the problem on locations.}
	\label{fig:figure_rec3_example_rec}
\end{figure}

\begin{figure}[!h]
	\centering
	\includegraphics[width=\linewidth]{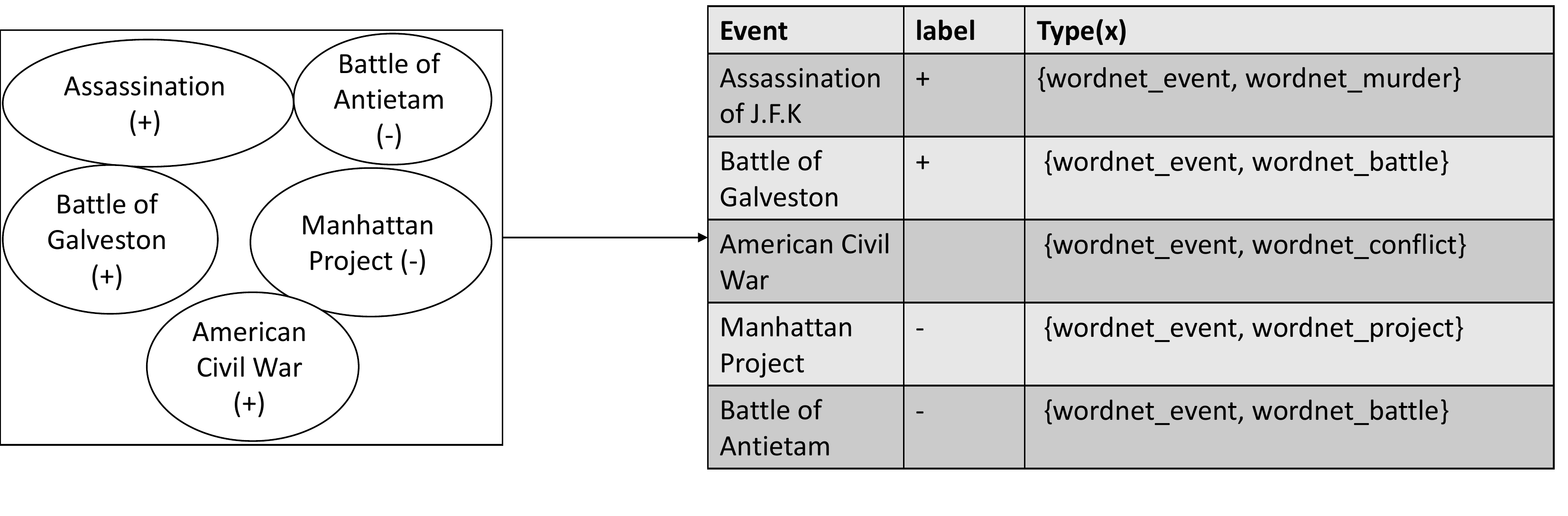}
	\caption{Construction of a recursive learning problem based on the ``Happened in" relation. Once the example set has been created, applicable relations are used to create a feature set for the newly constructed induction problem.}
	\label{fig:figure_rec3_problem_rec}
\end{figure}

\begin{figure}[!h]
	\centering
	\includegraphics[width=0.7\linewidth]{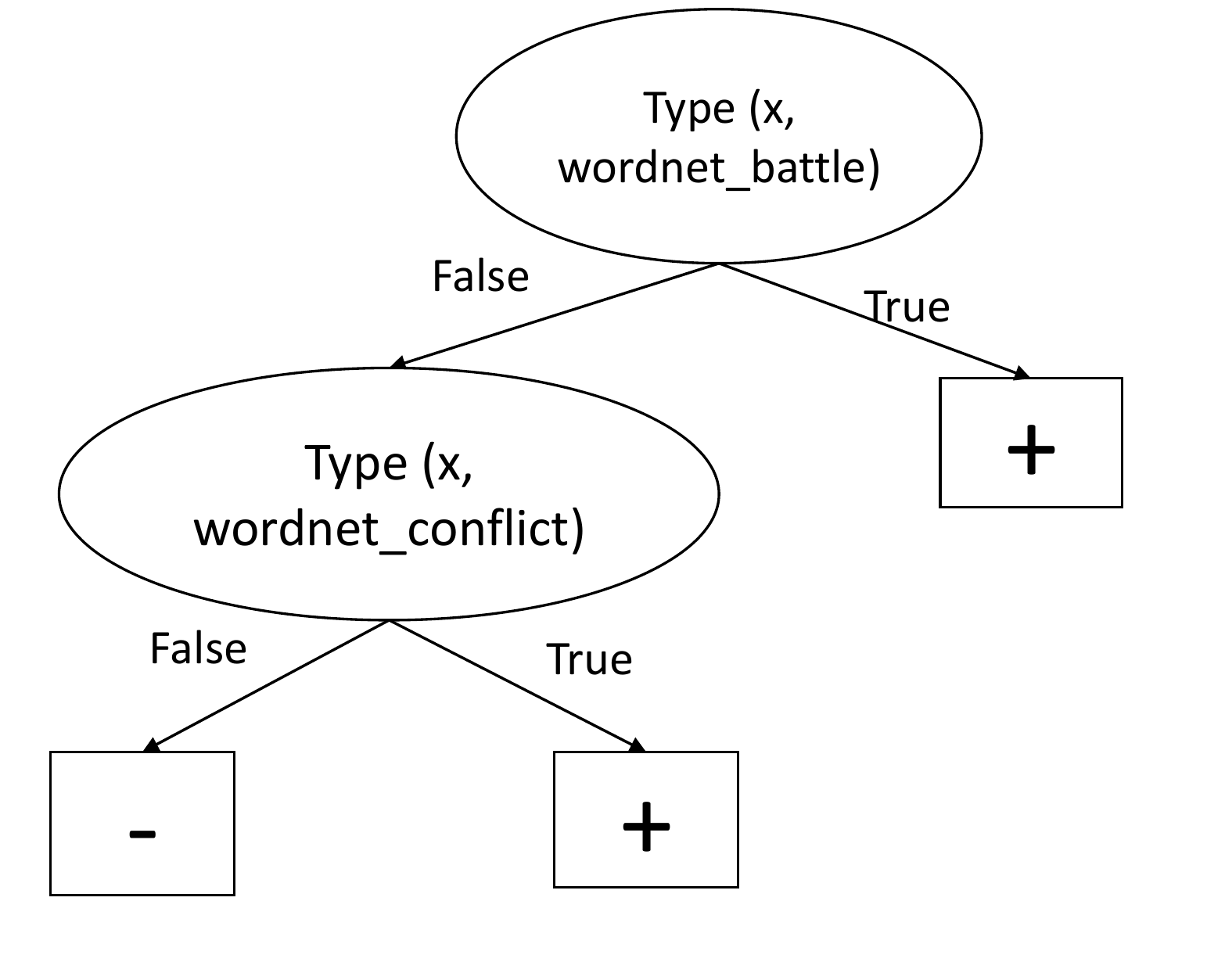}
	\caption{Recursive feature constructed by \emph{FEAGURE} for entities in the ``happened in" relation. This feature operates on events, and can be used in a classifier on locations.}
	\label{fig:figure_rec3_feature}
\end{figure}

Since we chose $d=2$ as the recursion depth parameter, the algorithm calls \emph{FEAGURE} recursively to try and generate new features for the new induction problem.
The values of the feature ``Happened in" are events. These events are used as objects for a recursive learning problem (Figure \ref{fig:figure_rec3_example_rec}).
We use the ``Type" relation as a feature, relying on hypernyms to classify events (Figure \ref{fig:figure_rec3_problem_rec}).
The resulting classifier (a decision tree induction algorithm was used) is shown in Figure \ref{fig:figure_rec3_feature}, and can be interpreted as ``is this event a battle or conflict?".

\begin{figure}[!h]
	\centering
	\includegraphics[width=\linewidth]{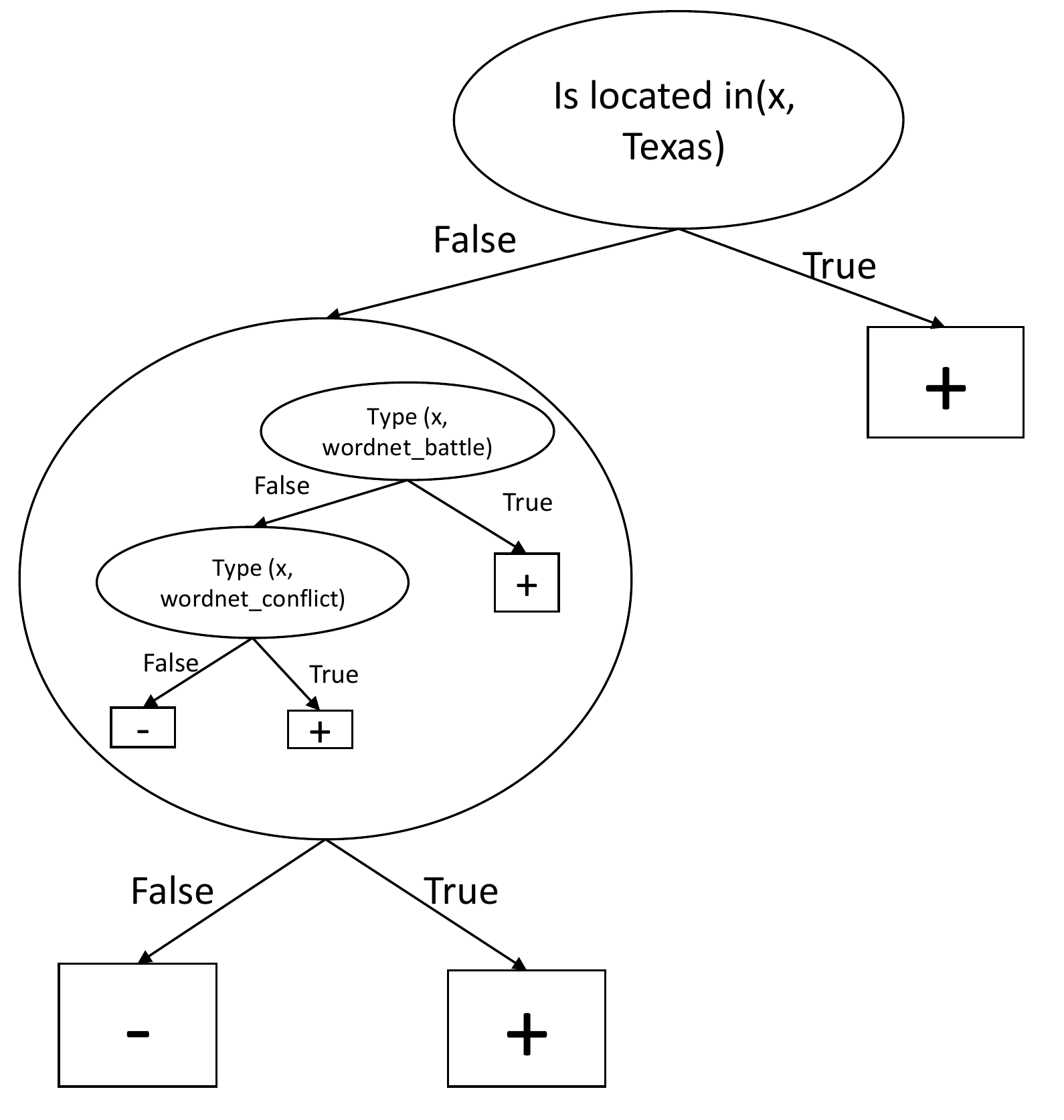}
	\caption{Final generated feature constructed by \emph{FEAGURE} for entities in the ``located in" relation. The feature in the left branch is the recursive feature constructed by applying \emph{FEAGURE} to the new learning problem.}
	\label{fig:figure_rec3_feature_full}
\end{figure}

Once we have generated this classifier on events, we can use it as a binary feature. \emph{FEAGURE} uses this new feature to expand the constructed induction problem on locations, shown in Figure \ref{fig:figure_rec3_problem}.
This feature is applied to a location through a majority vote over events that happened in that location. The result is a feature for locations representing the concept ``were most notable events in this location battles/conflicts?"
Finally, a decision tree learner is used on the expanded feature set to learn a classifier on locations to be used as a feature for our original learning problem. The new classifier for locations is shown in Figure \ref{fig:figure_rec3_feature_full}. It can be described as ``is this location located in Texas, or the site of battles or conflicts?".
Texts mentioning locations in and around Texas are more likely to link to locations that correspond to the output of this classifier.
We note that this feature was generated by \emph{FEAGURE}, and was later used by our external induction algorithm due to its high information gain. 

\section{Related work}

Many feature generation methodologies have been developed to search for new features that better represent the target concepts. There are three major approaches for feature generation: tailored methods, combinational techniques, and algorithms utilizing external knowledge.

Tailored approaches \cite{sutton1991learning,hirsh1994bootstrapping} are designed for specific problem domains and rely on domain-specific techniques. 
One such example is the bootstrapping algorithm \cite{hirsh1994bootstrapping}, designed for the domain of molecular biology. The algorithm represents features as nucleotide sequences whose structure is determined by existing background knowledge. The algorithm uses an initial set of feature sequences, produced by human experts, and uses a domain-specific set of operators to change them into new sequence features. 
Such special-purpose algorithms  may be effectively tailored for a given domain, but have proven difficult to generalize to other domains and problems.

Combinational feature generation techniques are domain-independent methods for constructing
new features by combining existing features. The LDMT algorithm \cite{utgo1991linear} performs feature construction in the course of building a decision-tree classifier. At each created tree node, the algorithm constructs a hyperplane feature through linear combinations of existing features in a way likely to produce concise, relevant hyperplanes. The LFC algorithm \cite{ragavan1993complex} combines binary features through the use of logical operators such as $\land ,\lnot$.
The FICUS algorithm \cite{markovitch2002feature} allows the use of any combinational feature generation technique, based on a given set of constructor functions. Recent work by \shortciteA{katz2016explorekit} uses a similar approach.

\emph{Deep Learning} \shortcite{rumelhart1986learning,lecun1998gradient} is another major class of combinational feature generation approaches. Here, the activation functions of the nodes can be viewed as feature schemes, which are instantiated during the learning process by changing the weights.

One limitation of combinational approaches is that they merely combine existing features to make the representation more suitable for the 
learning algorithm. 
Our \emph{FEAGURE} algorithm belongs to a third class of approaches that inject additional knowledge into the existing problem through the feature generation process.

Propositionalization approaches \shortcite{kramer2000bottom,cheng2011automatedfull} rely on relational data to serve as external knowledge. They use several operators to create first-order logic predicates connecting existing data and relational knowledge. 
\citeA{cheng2011automatedfull} devised a generic propositionalization framework  using linked data via relation-based queries. 
FeGeLOD \cite{paulheim2012unsupervisedfull} also uses linked data to automatically enrich existing data. 
FeGeLOD uses feature values as entities and adds related knowledge to the example,
thus creating additional features.

Unsupervised approaches allow us to utilize external knowledge, but they have a major issue:
Should we try to construct deep connections and relationships within the knowledge base, we would
experience an exponential increase in the number of generated features.
To that end, \emph{FEAGURE} and other supervised approaches use the presence of labeled examples to better generate deeper features.
Most supervised methods can trace their source to \emph{Inductive Logic Programming (ILP)} \cite{muggleton1991inductive,quinlan1990learning}, a supervised approach that induces a set of first-order logical formulae to separate the different categories of examples in the training set.
ILP methods start from single relation formulae and add additional relational constraints using the knowledge base, until formulae that separate the training set into positive and negative examples are found. To that end, these approaches make use of a \emph{refinement operator}. When applied on a relational formula, this operator creates a more specialized case of that formula. For example, given the logical formula $BornIn(X,Y)$, where $X$ is a person and $Y$ is a city, one possible refinement is the formula $BornIn(X,Y)\land CapitalOf(Y,Z)$, where $Z$ is a country. The result is a logical formula that considers a more specific case. Additionally, we can look at a refinement that restricts by a constant, turning $BornIn(X,Y)$ into, for example, $BornIn(X, \mbox{{United States}})$. This refinement process continues until a sufficient set of consistent formulae is found. 

An algorithm suggested by \citeA{terziev2011feature} shows an interesting approach to supervised feature generation. In his paper, \citeA{terziev2011feature} suggests a decision tree based approach, where in each node of the tree, an expansion of features is done similarly to FeGeLOD, with an entropy-based criterion to decide whether further expansion is required. This technique bears several similarities to \emph{Deep-FEAGURE} (Algorithm \ref{code-tree-thing}). Unlike \emph{Deep-FEAGURE}, the feature expansion
process is unsupervised, and the resulting feature must be a decision tree, restricting the generality
of the approach.

The \emph{dynamic feature generation} approach used by the SGLR algorithm \cite{popescul200716} can be seen as the supervised equivalent of propositionalization methods. Feature generation
is performed during the training phase, allowing for complex features to be considered by
performing a best-first search on possible candidates. This process allows SGLR to narrow the
exponential size of the feature space to a manageable number of candidates. While this supervised
approach overcomes the exponential increase in features that unsupervised approaches suffer
from, the space of generated features that it searches is significantly less expressive than that of our
approach. Through the use of recursive induction algorithm, our approach automatically locates
relationships and combinations that we would not consider otherwise.

Since we have focused on the problem of text classification, we also discuss a few text-based approaches for feature generation. 
Linguistic methods such as those described in a study by \shortciteA{moschitti2004complex} attempt to use part-of-speech and grammar information to generate more indicative features than the bag-of-words representation of texts.
Concept-based approaches are the most similar to our own. Two examples of such methods are Explicit Semantic Analysis (ESA) \cite{gabrilovich2009wikipediafull},  which generates explicit concepts from Wikipedia based on similarity scores, and Word2Vec \shortcite{mikolov2013distributed}, which generates latent concepts based on a large corpus. Both approaches provide us with a feature set of semantic concepts. These concepts can be used alongside our approach, allowing us to induce over them.

\section{Conclusions}

When humans use inductive reasoning to draw conclusions from their experience, they use a vast amount of general and specific knowledge. In this paper we introduced a novel methodology for enhancing existing learning algorithms with background knowledge represented by relational knowledge bases.
The algorithm works by generating complex features induced using the available knowledge base. It does so through the extraction of recursive learning problems based on existing features and the knowledge base, that are then given as input to induction algorithms. The output of this process is a collection of classifiers that are then turned into features for the original induction problem.

An important strength of our approach is its generality. The features generated by \emph{FEAGURE}
can be used by any induction algorithm, allowing us to inject external knowledge in a general,
algorithm independent manner. One potential limitation of our approach is that it requires features
with meaningful values in order to operate. Despite this limitation, there is a wide range of problems
where feature values have meaning. For these problems, we can apply one of several general and
domain-specific knowledge bases, depending on the problem.
In recent years, we have seen an increase in the number of available knowledge bases. These
knowledge bases include both general knowledge bases such as Freebase, YAGO, Wikidata and the
Google Knowledge Graph, and more domain-specific knowledge bases, from the British Geographical
Survey (BGS) linked dataset, containing roughly one million geological facts regarding various
geographical formations, through biological databases such as Proteopedia, composed of thousands
of pages regarding biological proteins and molecules, to entertainment-focused databases such as
IMDB, containing millions of facts on movies, TV-series and known figures in the entertainment
industry. With the recent surge of well-formed relational knowledge bases, and the increase in use
of strong learning algorithms for a wide variety of tasks, we believe our approach can take the
performance of existing machine learning techniques to the next level.

\clearpage
\bibliographystyle{theapa}
\bibliography{document}

\end{document}